\documentclass[conference]{IEEEtran}
\IEEEoverridecommandlockouts
\usepackage{adjustbox}
\usepackage{cite}
\usepackage{amsmath,amssymb,amsfonts}
\usepackage{algorithmic}
\usepackage{graphicx}
\usepackage{textcomp}
\usepackage{xcolor}
\usepackage{multirow}
\usepackage{hyperref}
\usepackage{booktabs}
\usepackage{hhline}
\usepackage{amssymb}
\usepackage{soul}
\usepackage{fancyhdr}   

\usepackage{fancyhdr}
\def\BibTeX{{\rm B\kern-.05em{\sc i\kern-.025em b}\kern-.08em
    T\kern-.1667em\lower.7ex\hbox{E}\kern-.125emX}}
    
\def\etal{\emph{et al}.}
    
\definecolor{rubinered}{HTML}{CE0058}
\newcommand{\thomas}[1]{\textcolor{rubinered}{[#1 -- TP]}}
\definecolor{green}{rgb}{0.0, 0.65, 0.31}

\definecolor{bleudefrance}{rgb}{0.19, 0.55, 0.91}

\definecolor{ao(english)}{rgb}{0.0, 0.5, 0.0}

\definecolor{violet}{HTML}{6a51a3}

\begin{document}

\title{Fine-grained Human Activity Recognition Using Virtual On-body Acceleration Data
}

\makeatletter
\newcommand{\linebreakand}{%
  \end{@IEEEauthorhalign}
  \hfill\mbox{}\par
  \mbox{}\hfill\begin{@IEEEauthorhalign}
}
\makeatother

\author{}

\author{\IEEEauthorblockN{Zikang Leng}
\IEEEauthorblockA{\textit{College of Computing} \\
\textit{Georgia Institute of Technology}\\
Atlanta, USA \\
zleng7@gatech.edu}
\and
\IEEEauthorblockN{Yash Jain}
\IEEEauthorblockA{\textit{School of Interactive Computing} \\
\textit{Georgia Institute of Technology}\\
Atlanta, USA \\
yashjain@gatech.edu}
\and
\IEEEauthorblockN{Hyeokhyen Kwon}
\IEEEauthorblockA{\textit{Department of Biomedical Informatics} \\
\textit{Emory University}\\
Atlanta, USA \\
hyeokhyen.kwon@dbmi.emory.edu}
\linebreakand
\IEEEauthorblockN{Thomas Plötz}
\IEEEauthorblockA{\textit{School of Interactive Computing} \\
\textit{Georgia Institute of Technology}\\
Atlanta, USA \\
thomas.ploetz@gatech.edu}
}

\maketitle

 \thispagestyle{fancy}
\fancyhead{} 
\fancyfoot{} 
\pagenumbering{gobble}
\fancyfoot[C]{\textcolor{red}{This manuscript is under review. Please write to zleng7@gatech.edu for up-to-date information}}

\begin{abstract}
Previous work has demonstrated that virtual accelerometry data, extracted from videos using cross-modality transfer approaches like IMUTube, is beneficial for training complex and effective human activity recognition (HAR) models.
Systems like IMUTube were originally designed to cover activities that are based on substantial body (part) movements.
Yet, life is complex, and a range of activities of daily living is based on only rather subtle movements, which bears the question to what extent systems like IMUTube are of value also for fine-grained HAR, i.e., \textit{When does IMUTube break?}
In this work we first introduce a measure to quantitatively assess the subtlety of human movements that are underlying activities of interest--the motion subtlety index (MSI)--which captures local pixel movements and pose changes in the vicinity of target virtual sensor locations, and correlate it to the eventual activity recognition accuracy.
We then perform a ``stress-test'' on IMUTube and explore for which activities with underlying subtle movements a cross-modality transfer approach works, and for which not.
As such, the work presented in this paper allows us to map out the landscape for IMUTube applications in practical scenarios.

\end{abstract}


\section{Introduction}




The effectiveness of supervised learning methods for deriving human activity recognition systems (HAR) for wearables depends heavily on the availability of curated, i.e., annotated datasets \cite{chen2021sensecollect}.
One major issue with current machine learning solutions in the field is the paucity of labeled datasets.
Annotating sensor data in HAR is expensive, often privacy invasive, and often prone to errors or has other practical limitations \cite{kwon2019handling,jiang2021research,cilliers2020wearable}. 

Recently, systems like IMUTube \cite{kwon2020imutube} have been introduced that tackle the aforementioned problem by generating virtual IMU data from videos 
to increase the size of training datasets that can be used for model training.
IMUTube was designed to recognize substantial body movements through estimating the movements of 3D body keypoints from single-view videos.
The system has been successfully validated with virtual IMU data for wrist sensors for locomotion and gym exercise activities, i.e., targeting activities that are based on substantial body movements (or of parts thereof).

\begin{figure*}[t]
    \centering
    \includegraphics[width=.95\linewidth]{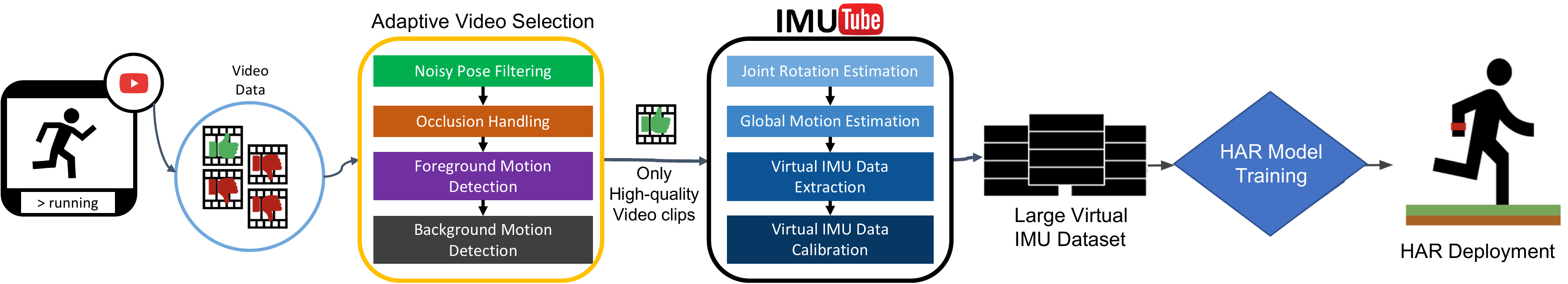}
    \caption{IMUTube system that generates virtual IMU data from unconstrained videos of human activities \cite{kwon2021approaching}.
    Video data is queried through activity keyword from public repositories, such as YouTube. 
    An Adaptive Video Selection module filters out video segments that bear high risk of failing the 3D motion tracking due to noisy poses, occlusions, or foreground and background motion blurs.
    The rest of the videos is used to estimate 3D motion information from ongoing human activities, where virtual IMU data is extracted and calibrated for target real IMU sensor for deployment.
    Once virtual IMU data have been generated, HAR models can be trained based on the augmented training dataset with weak labels in form of the original search query keyword or video title and descriptions.
    Then, the trained model is deployed in the wild.
    (Figure adopted from Kwon \etal \cite{kwon2021complex} and used with permission). 
    }
    \label{fig:imutube}
\end{figure*}

With the previous success of systems like IMUTube (Fig.\ \ref{fig:imutube}) for recognizing human activities with underlying coarse motions, the next step now is to explore to what extent such approaches generalize to activities with more subtle body movements.
Many important activities of daily living are more subtle, exhibiting much lesser motion differences between and more variance within activities, e.g.,
eating, holding (and talking on) a phone, or washing dishes, etc~\cite{thomaz2015practical}.
As such, it is desirable to understand how substantial the body (part) movements need to be for systems like IMUTube to be of practical value.

The work presented in this paper targets exactly that aforementioned problem:
We develop a measure for the quantitative assessment of the subtlety of movements and correlate it to the effectiveness of HAR systems that have been derived using virtual IMU training data generated using IMUTube on unconstrained videos of target activities.
As such, we aim to find the ``breaking point'' of IMUTube with respect to when the motions assigned with activities of interest are too subtle for the system to be beneficial for designing HAR systems.
This analysis allows us to formulate concrete guidelines for HAR practitioners who plan to employ cross-modality transfer systems for their HAR model development efforts.



We hypothesize that the amount of human motion in a video sequence can be captured by measuring local pixel movements in the vicinity of the target on-body sensor locations.
We define a novel metric--the \textit{Motion Subtlety Index} (MSI)--that measures the subtlety of motion of human activities performed in a video sequence with respect to the particular on-body sensor locations of interest by using optical flow and pose estimation methods.
Our experimental evaluation on a range of activities of daily living shows that the MSI extracted from human activity videos is highly correlated to the 
the eventual recognition accuracy of HAR systems that were derived using virtual IMU data extracted from videos. 
As such, the MSI is an excellent proxy that can be used for the \textit{a-priori} prediction of the potential effectiveness of cross-modal transfer approaches.

With this new measure, we are now in the position to ``stress-test'' the IMUTube system.
We study its effectiveness for virtual wrist sensors on a much more extensive range than in the original publications when the system was first introduced.
We focus on activities with more subtle motions than the coarse motions studied previously, including .
washing hands, playing instruments, driving, and so on, which are essential activities when it comes to assessing an individuals' quality of life \cite{inproceedings}.

Our experiment shows that activities with more subtle motions can benefit from virtual IMU data extracted from videos.
In the second part of our study we focus specifically on an eating detection task \cite{thomaz2015practical} based on wrist-worn IMUs, which has important applications in health assessments \cite{morshed2022food,konttinen2019depression,navarro2021effects,van2018causes}.
On this task, IMUTube showed surprising effectiveness for improving recognition accurcy with gains of 8.4\% absolute when using curated video data and of 5.9\% absolute (both F1-scores) when using unconstrained video data as they were retrieved through a keyword-based search on YouTube.
Such performance improvements are significant because IMUTube was designed with coarse, i.e., significant body motion in mind, contrasting with the subtle movements in daily activities.

The contributions of this paper are two-fold:
\begin{enumerate}
\item By correlating activity recognition accuracy, measured in F1-score, and MSI, our study allows us to identify when IMUTube ``reaks''. i.e., to assess the application range of cross-modality transfer approaches for developing HAR systems.

\item Through the newly introduced quantification of the subtlety of human movements and its correlation to the eventual effectiveness of HAR systems that were derived based on virtual IMU data, we can draw conclusions about application cases for systems like IMUTube and, essentially, map out the landscape for IMUTube applications in practical scenarios.
\end{enumerate}

Our results shall encourage the HAR community to investigate further the use of cross-modality transfer approaches such as IMUTube for various complex and subtle human activities.




\section{Related Work}

In this work, we analyze to what extent virtual IMU data from wrist sensors are of value for deriving recognition systems that target activities with underlying subtle motions.
In what follows, we briefly explain the virtual sensor generation method through IMUTube, on which our work is based, before summarizing previous work in classifying fine motion activities in daily life. 

\subsection{IMUTube: Virtual IMU Data from Videos}

IMUTube was proposed to tackle the challenges that come with the collection of labeled wearable data such as privacy, accuracy, but also practical and logistical obstacles.
The system automatically converts large-scale video datasets into virtual IMU data that can be used for training sensor-based HAR systems \cite{kwon2020imutube, kwon2021approaching,kwon2021complex}.
As shown in \autoref{fig:imutube}, IMUTube resembles a processing pipeline that incorporates computer vision, graphics, and machine learning models to extract virtual IMU data from human activity videos for various on-body locations. 
The system consists of three main parts: 
\textit{i)} adaptive video selection;
\textit{ii)} 3D human motion tracking; and 
\textit{iii)} virtual IMU data extraction and calibration.

When querying YouTube with activity keywords, for example, eating, the returned videos may contain: 
\textit{i)} many sequences that are irrelevant to human activities, such as intros and outros; or 
\textit{ii)} video sequences that are very challenging to track human motions accurately. 
To identify video segments that will lead to high-quality virtual IMU data, the adaptive video selection module actively filters video sequences with irrelevant frames, noisy poses, occlusions, or extreme foreground and background camera motions.

The remainder of the videos selected from this process is then processed to estimate 3D joint rotation and global motion for extracting 3D motion information of human activities. 
Virtual IMU data is then extracted using the 3D motion information.
The extracted virtual IMU data is calibrated using a small amount of target real IMU data to overcome the inevitable gap between the distributional characteristics of the two domains.
IMUTube was mainly designed, deployed and validated for human activities with underlying coarse motions, such as locomotion or gym exercises.
Yet, many real-world activities are based on only subtle body (part) movements and, hence, in this paper we explore how far the application range of approaches like IMUTube reaches.

\subsection{Sensor-Based Recognition of Daily Activities}

To assess well-being or quality of life for individuals, understanding activities in daily routines is important. \cite{kushner2013lifestyle,phillips2020lifestyle}
For example, to assess the--potentially sedentary--lifestyle of individuals, early works in human activity recognition with wearable sensors (HAR) focused on detecting locomotion activities, including sitting or walking \cite{stisen2015smart,sztyler2016body,zhang2012usc}.
After achieving success in recognizing locomotion activities, more recent work explored the application of wearable-based human activity recognition to fine-grained daily activities that provide more context for how users are situated.

Due to the popularity of commodity wrist devices, many works have studied fine-grained activity recognition with wrist sensors.
Laput and Harrison~\cite{laput2019sensing} demonstrated that fine-grained hand activities, such as writing or typing on a keyboard, can be detected by accelerometry data from wrist motions.
Moschetti~\etal~\cite{moschetti2017daily} used wrist and index finger motion data to recognize eating, drinking, and brushing teeth.
Ashri~\etal~\cite{inproceedings} have explored the use of wrist motion sensors to recognize daily activities such as driving, getting dressed, or house cleaning.
Other work showed that wrist or arm sensors could successfully classify multiple gym exercises~\cite{morris2014recofit,koskimaki2017myogym,koskimaki2014recognizing,um2017exercise}.

With the increasing attention to eating activities for measuring risks in physical and mental health~\cite{konttinen2019depression,navarro2021effects,van2018causes}, some work has focused on designing HAR models that are specific to eating detection using werables~\cite{thomaz2015practical,bedri2017earbit,bedri2020fitbyte,bedri2015wearable}.
Eating as an activity is a very subtle motion. 
It involves repeated hand movement to the mouth, which varies vastly depending on the eating habits and the food consumed. 
One of the critical challenges in a fully automated food intake monitoring system is to identify whether the eating has happened or not within a given time frame. 
Past works have tried to solve this problem using training an eating moment recognition method on carefully curated in-lab datasets and then testing upon semi-controlled wild datasets\cite{thomaz2015practical,morshed2020real}.

While previous methods for recognizing daily activities were shown to be effective, they are still limited by the amount of lab data available for training.
For such activities with subtle motions, collecting large-scale labeled datasets is pivotal for improving the generalizability of the trained model.


\section{Quantifying Motion Subtlety in Human Activity Video}

The IMUTube system was originally designed to capture substantial body (part) movements in human activity videos -- information, which is subsequently transferred to \textit{virtual} IMU data.
In this work, we explore how such a system can be used when studying human activities with only \textit{subtle} underlying movements.
Essentially, we aim to find the ``breaking point'' of IMUTube-like approaches, i.e., what the limits of its applicability are -- beyond the originally studied coarse movement activities.
We define subtle motions in human activities as those movements that involve one or two body parts moving in the very limited range of distances.
For example, activities such as  washing hands, writing, cooking, eating, and so on, involve only hand or arm movements that are much smaller in range compared to activities that are based on whole-body involvement, such as sports or gym exercises.

\begin{figure}
    \centering
    \includegraphics[width=.95\linewidth]{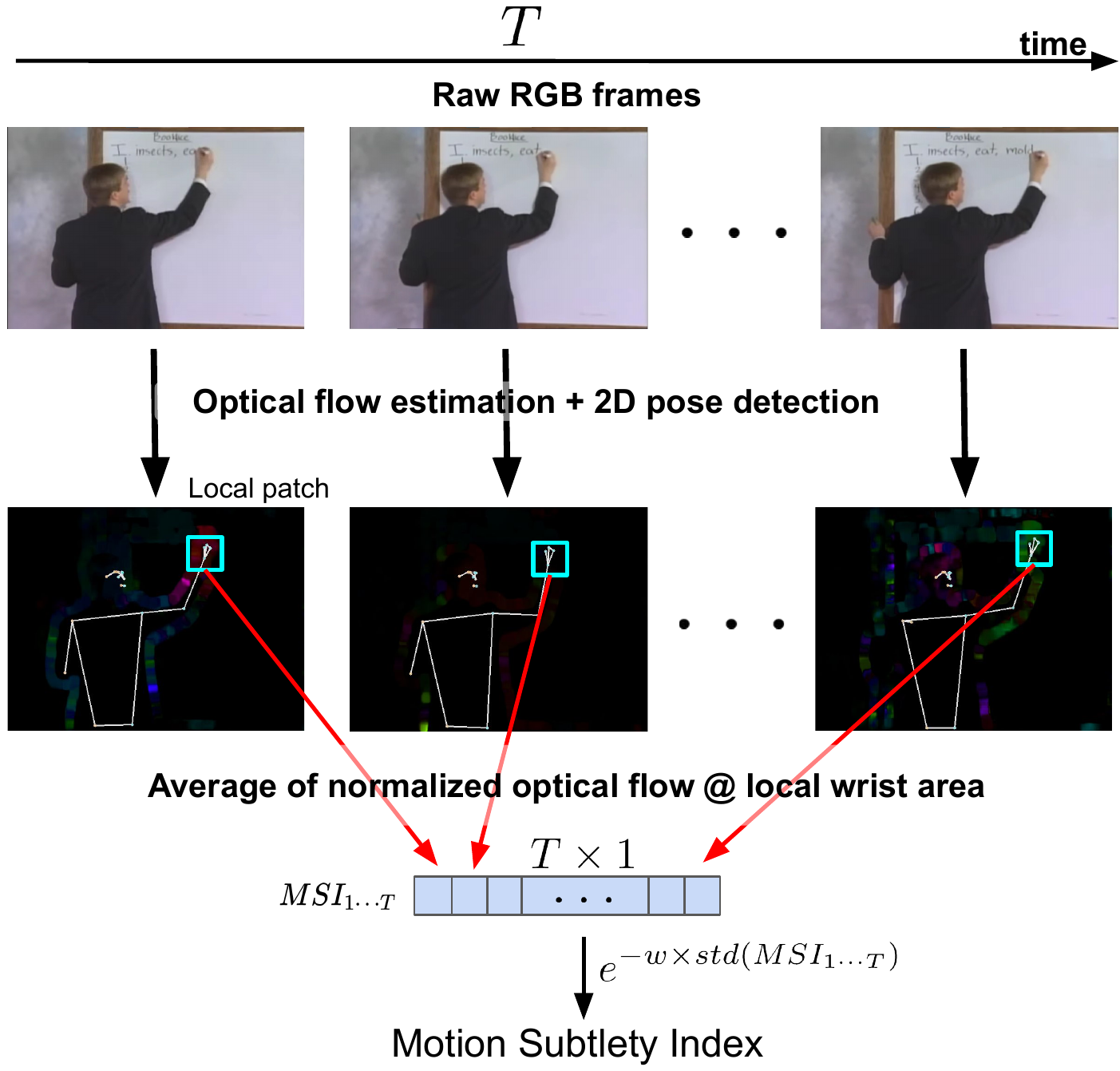}
    \vspace{-0.0in}
    \caption{Motion Subtlety Index (MSI) calculated from video frames. 
    From the analysis frame with window size $T$, we first estimate optical flow and 2D pose from each frame.
    The optical flow is normalized according to the size of frame.
    From each frame, we average the normalized optical flow from the local on-body sensor location, $MSI_{1\cdots T}$ (wrist in this example).
    For the local region patch, we use a square patch of size $K=0.02 \times max(H, W)$, where $H$ and $W$ are height and width of the frame size, respectively, to take account of varying frame size across videos.
    Then, MSI is computed as a exponential of the negative standard deviation of $MSI_{1\cdots T}$, where smaller MSI value indicates more motions in the observed activity sequence.
    }
    \label{fig:msi}
\end{figure}


3D human motion tracking is at the heart of IMUTube.
Although state-of-the-art 3D human motion tracking techniques are effective at capturing whole body movements, they are still limited at capturing subtle movements that only involve local body parts \cite{desmarais2021review}.
We consider the IMUTube system as upper bounded by the amount of motion that motion tracking techniques can capture in videos of  human activities.
Therefore, we hypothesize that IMUTube will start to struggle to generate useful virtual IMU data as the motions involved in underlying activity become more subtle.

With this in mind, it is important to identify early on for which activities IMUTube may ``break'' before actually allocating resources and time to generate virtual IMU data and develop activity recognition model~\cite{hiremath2020deriving}. 
For an objective quantification of the subtlety of movements involved in an activity of interest, we introduce a novel metric, namely \textit{Motion Subtlety Index} (MSI).
This metric measures the motion subtlety of activities from videos, which supports the aforementioned early decision making whether using IMUTube is beneficial or not.
Specifically, we combine optical flow \cite{zhai2021optical}
 and 2D pose estimation \cite{dubey2022comprehensive} from a video sequence to quantify the pixel-level motion quantity in the local area near the targeted on-body sensor location.

\autoref{fig:msi} illustrates how the MSI is calculated for an exemplary video segment that captures a sequence of a writing activity with the wrist as the target location for the virtual IMU sensor. 
For each frame in a video segment with $T$ frames, we first compute the optical flow and estimate 2D poses. 
The estimated optical flow at each pixel and time is normalized according to frame size to take account for different resolution of videos available, $(u^t_i, v^t_i) \rightarrow (u^t_i/H, v^t_i/W)$, where $(u^t_i, v^t_i)$ are vertical and horizontal optical flow at pixel $i$ and time $t$ and $(H, W)$ are height and width of frame size of the video.
Next, at each frame, we calculate the average magnitude of the normalized optical flow at the local patch, $K\times K$, in the neighborhood of the wrist keypoint location, which is automatically detected by our 2D pose estimation procedure~\cite{fang2017rmpe}.
To take account of varying resolution of video frames, the patch size is 2\%  of the larger dimension of the frame, $K=0.02\times max(H, W)$.
\begin{equation}
    MSI_t = \frac{1}{N} \sum_{-\frac{K}{2}\geq i,j \geq \frac{K}{2}}  \sqrt{(u^t_{n + i})^2 + (v^t_{n + j})^2},
\end{equation}
where $u^t_n$ and $v^t_n$ are vertical and horizontal components of the normalized optical flow measurements from the keypoint location at time $t$, and $N=K \times K$.
The MSI for the analysis window is then computed as the exponential of negative standard deviation of $MSI_{1\cdots T}=[MSI_1, MSI_2, \cdots, MSI_T]$ :

\begin{equation}
    MSI = e^{-w\times std(MSI_{1\cdots T})}  
\end{equation}

We used $w=100$ to account for the minimal difference in MSI where $std(MSI_{1\cdots T}) \approx 0$.

Overall, MSI captures the motion information recorded around the on-body sensor location for underlying activities in the given video sequence.
A smaller MSI means more significant motions are involved with ongoing activities, whereas a larger MSI indicates more subtle movements. Note that computing the MSI for any given video segment takes far less time and resources than extracting the virtual IMU data from the video segment using IMUTube.

\section{Experiment}

In this study, we look at potential changes in the accuracy (F1 score) of human activity recognition models trained using virtual IMU data when activities with more subtle movements are taregted (quantified using our newly proposed MSI measure).
We first give an overview of our workflow that utilizes IMUTube.
Then we will discuss the benchmark datasets and video sources used for our study.

\subsection{HAR with Virtual IMU Data from Wrist Sensors}

We followed the approach used in the original IMUTube experiments \cite{kwon2020imutube}.
We extracted virtual IMU data for wrist sensors from video datasets, which are calibrated with (a small amount of) real IMU data used for training.
Real IMU data is subsampled to 25 Hz to match the sampling rate of virtual IMU data extracted from videos.
Both virtual and real IMU data are segmented into fixed length analysis frames that contain consecutive readings using the sliding window technique. 
The choice of window size depends on the particular classification task of interest.
Then, features are extracted from both real and virtual analysis windows, which are used to train a classification back-end.
In line with previous IMUTube studies, we employ the Random Forest classifier.
Lastly, the trained model is tested on unseen real IMU data.
For assessing motion subtlety, MSI is derived from the video sequence that corresponds to each virtual IMU data segment.

As baseline, we use the model that was trained only using real IMU data, i.e., the training sets as they were provided by the individual datasets, which is in line with previous work. 
Identical to previous IMUTube explorations, we then train a model using both real and virtual IMU data, which is used to assess the effectiveness of virtual IMU data for HAR tasks.
Model effectiveness is evaluated through F1 scores.
As explained below, virtual IMU datasets are extracted from curated or in-the-wild video datasets.

\begin{table}
    \centering
    \caption{Dataset Statistics for eating and daily activity classification experiments. 
    For eating activity classification, real IMU data from lab (Lab-20) and in the wild (Wild-7) were taken from Thomaz \etal \cite{thomaz2015practical}.
    For daily activity classification, real IMU data were taken from HAD-AW \cite{inproceedings}.
    Curated and In-the-wild videos used for extracting virtual IMU data are taken from Kinetics-400~\cite{kay2017kinetics} and YouTube, respectively. 
    }
    \begin{tabular}{c|c|c}
         Task & Dataset & Duration \\
         \hline\hline
         \multirow{4}{*}{Eating} & Real IMU (Lab-20) & 284 minutes \\ 
         \cline{2-3}
         & Real IMU (Wild-7) & 128 minutes \\
         \cline{2-3}
         & Virtual IMU (Curated) & 17 minutes \\
         \cline{2-3}
         & Virtual IMU (In-the-wild) & 31 minutes \\
         \hline\hline
         \multirow{3}{*}{Daily Activity} & Real IMU (HAD-AW) & 454 minutes \\
         \cline{2-3}
         & Virtual IMU (Curated) & 201 minutes \\
         \cline{2-3}
         & Virtual IMU (In-the-wild) & 71 minutes \\
         \hline
    \end{tabular}
    \label{tab:dataset}
\end{table}

\subsection{Benchmark Real IMU Dataset}

For studying subtle motions in daily activities, we use two benchmark datasets for wrist sensors: daily activity classification~\cite{inproceedings} and eating activity classification~\cite{thomaz2015practical}.
\autoref{tab:dataset} shows statistics of the dataset used in our experiments.

\subsubsection{Daily Activities}

We used HAD-AW dataset \cite{inproceedings} as our benchmark dataset, which consists of 31 activity classes from subtle to coarse motions collected using an Apple Watch. 
Out of a total of 31 activity classes, we selected 17 activity classes that are based on more subtle movements of the body (parts):
\textit{
playing violin,
playing piano,
playing guitar,
driving automatic,
driving manual,
reading,
writing,
eating,
cutting components,
washing dishes,
washing hands,
showering,
sweeping,
wiping,
drawing,
flipping,
bed-making}.

\subsubsection{Eating}

We used the large public dataset from Thomaz \etal \cite{thomaz2015practical} that contains wrist-recorded IMU data from both in-lab and in-the-wild eating settings.
The \textit{Lab-20} dataset was collected from 21 participants in the lab covering both eating and non-eating activities for approximately 31 minutes period on average. 
The eating moments involve,
eating with fork \& knife, 
hand, 
and spoon.
While non-eating activities include 
watching trailer, conversation,
brush teeth,
place a phone call etc., all being subtle hand motions. 
For in-the-wild dataset, we used the \textit{Wild-7} dataset \cite{thomaz2015practical}, which was collected from seven participants. 
Out of a total of 31 hours and 28 minutes of data, 2 hours and 8 minutes were labeled as eating activities.

\subsection{Curated Video Data Set}

For assessing the feasibility of using virtual IMU data for recognizing subtle motion activities, we first collected virtual IMU data from a well curated video dataset, Kinetics-400 \cite{kay2017kinetics}, where all videos were manually trimmed down to clips of 10 seconds to contain specific activity classes. 

\subsubsection{Daily Activities}

\begin{figure}
    \centering
    \includegraphics[width=\columnwidth]{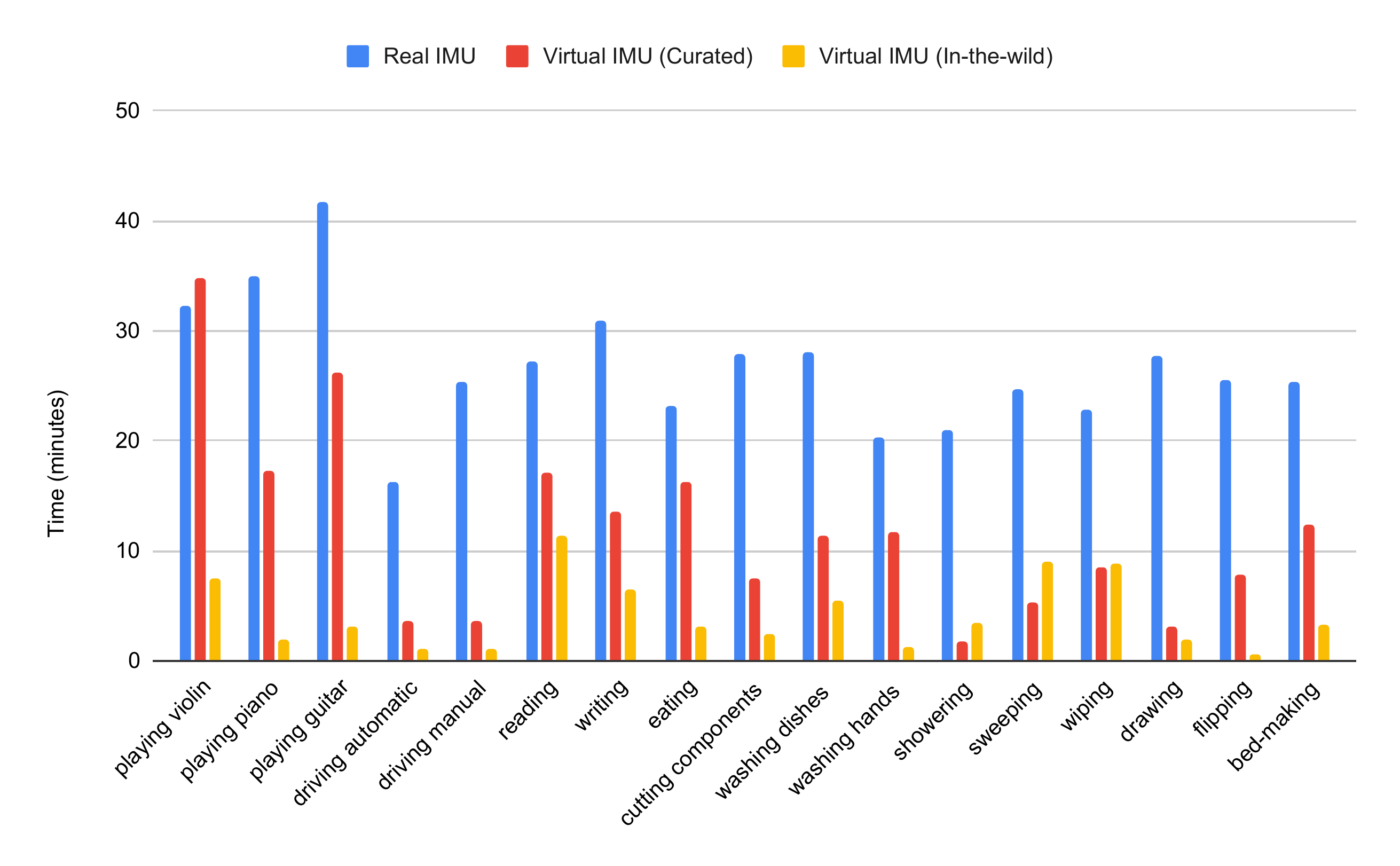}
    \caption{Comparison of sizes of real and virtual (curated and wild) IMU datasets for daily activity recognition experiments.
    The size of virtual IMU datasets is limited for some classes, e.g.,  driving automatic/manual, sweeping, etc., due to the lower quality of  videos containing such activities which hinders the generation of high-quality virtual IMU data (as determined by the IMUTube system). Examples: zoomed-in scenes for the hand movements, or viewing angle not showing wrist movements.}
    \label{fig:daily_activity_data_proportion}
\end{figure}

For the first set of experiments we used the well curated Kinetics-400 video dataset \cite{kay2017kinetics}, and focused on 17 activity classes that are based on subtle body (parts) movements (as assessed by us).
In total, we generated a virtual IMU dataset with approximately 50\% of the size of the corresponding, original real IMU dataset.
As shown in \autoref{fig:daily_activity_data_proportion}, the virtual IMU samples are imbalanced, as many video clips are removed due to low quality and noisy scenes for extracting virtual IMU for wrist movement. Hence, the duration of the extracted virtual IMU dataset is shorter than the combined duration of the video clips. 
\subsubsection{Eating}

From the Kinetics-400 dataset, we collected 417 video clips for the eating class, which were labeled as one of 10 eating-related classes (eating: burger, cake, carrots, chips, doughnuts, hotdog, ice cream, nachos, spaghetti, watermelon). 
The number of eating instances in the extracted virtual IMU dataset is approximately 6\% of that in the Lab-20 dataset. 

\subsection{In-the-Wild Video Dataset}

YouTube houses one of the largest collection of uncurated videos that are coarsely labeled using Google's search algorithm, providing a virtually unlimited source of videos for any activity. 
As shown in \autoref{fig:eating}, videos from such repositories may contain scenes that do not capture movements related to eating, i.e., scenes that IMUTube can automatically filter out with Adaptive Video Selection modules (yellow box;\autoref{fig:imutube}).
We downloaded videos from YouTube using activity names as search queries. Returned videos are sorted according to relevance.

\subsubsection{Daily Activities}

We downloaded one video for each class, which was on average approximately 8 minutes long (per activity).
Based on this, we generated virtual IMU data that are in total approximately 16\% the size of the baseline, real IMU dataset.

\subsubsection{Eating}

We downloaded two YouTube videos, where the first video is 22 minutes long, and the second one has a duration of 17 minutes.
From those two videos combined, we could collect 11\% of the amount of eating data samples compared to Lab-20 dataset.

\begin{figure}
    \centering
    \small
    \begin{tabular}{c c c c}
        \includegraphics[width=.2\linewidth]{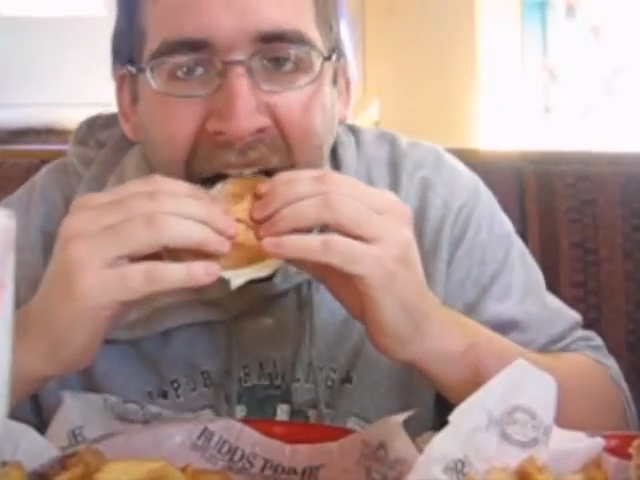} &
        \includegraphics[width=.2\linewidth]{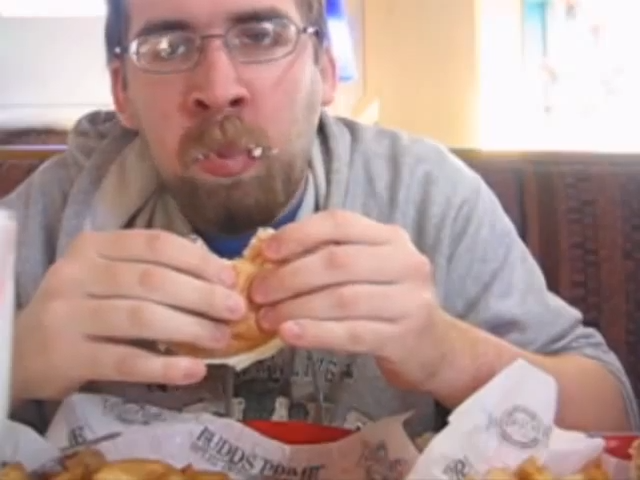} &
        \includegraphics[width=.2\linewidth]{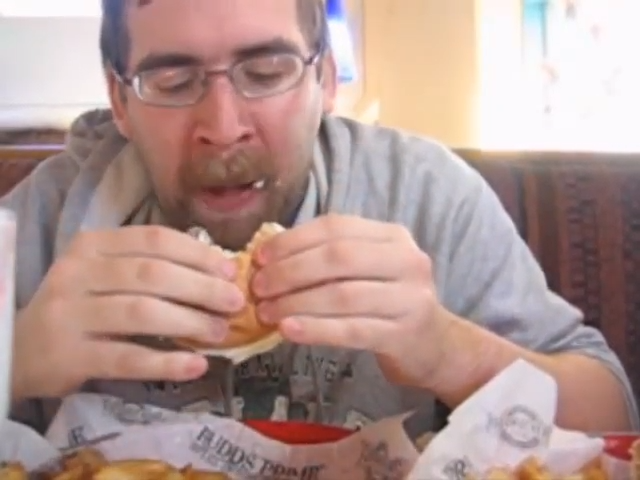} &
        \includegraphics[width=.2\linewidth]{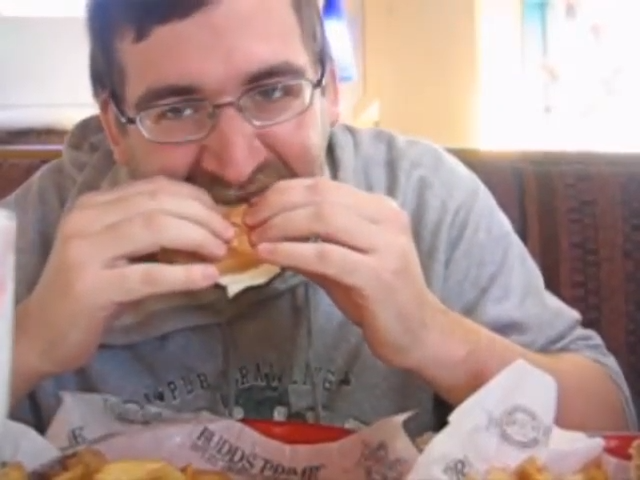} \\
        \includegraphics[width=.2\linewidth]{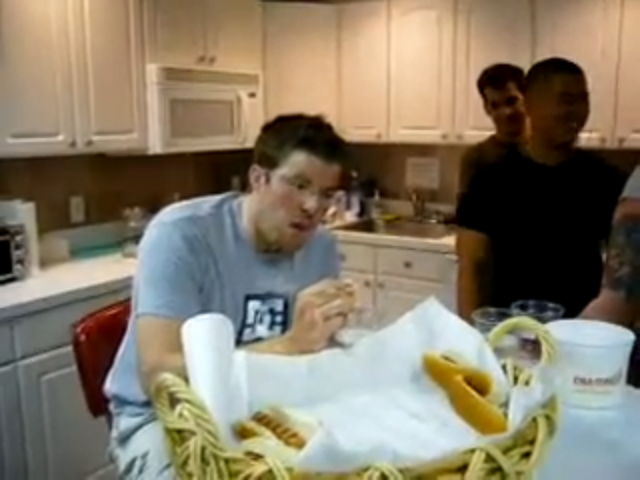} &
        \includegraphics[width=.2\linewidth]{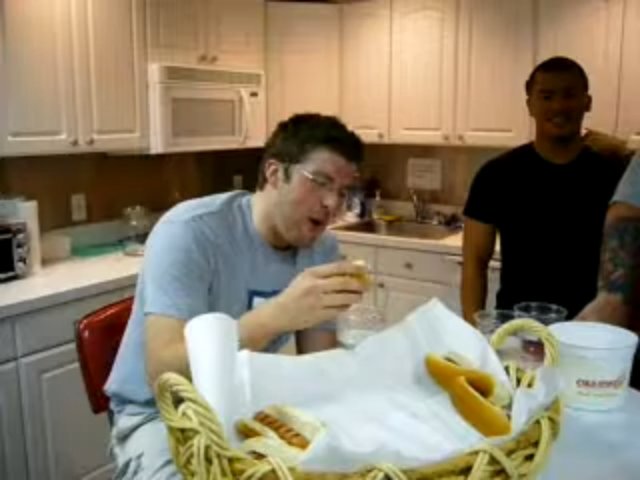} &
        \includegraphics[width=.2\linewidth]{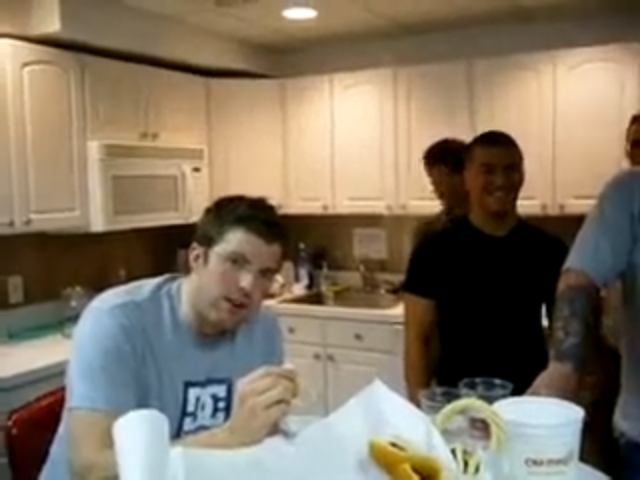} &
        \includegraphics[width=.2\linewidth]{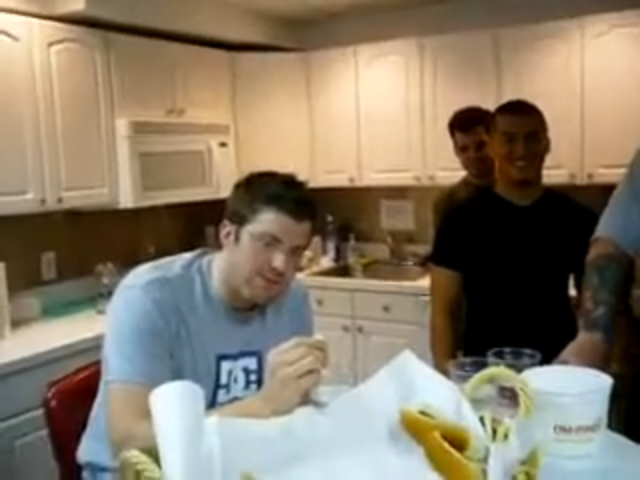}\\
        \multicolumn{4}{c}{(a) Curated video dataset (Kinetics-400)} \\
        \includegraphics[width=.2\linewidth]{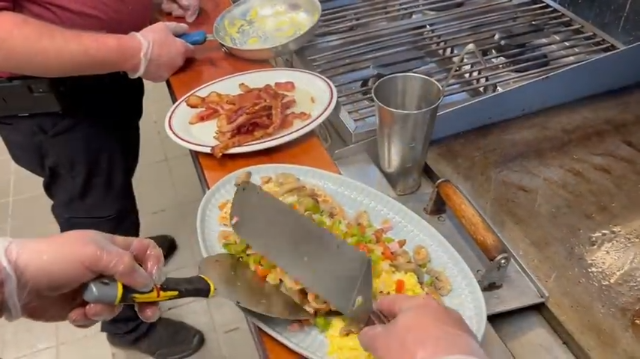} &
        \includegraphics[width=.2\linewidth]{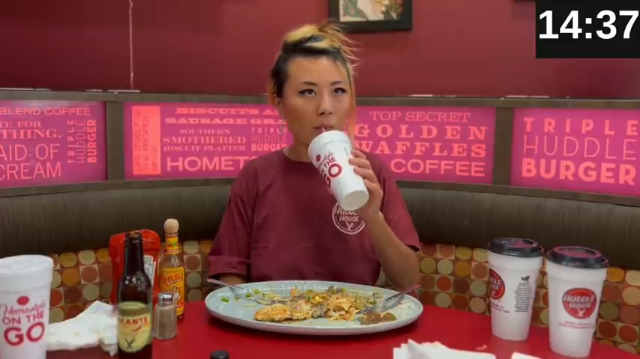} &
        \includegraphics[width=.2\linewidth]{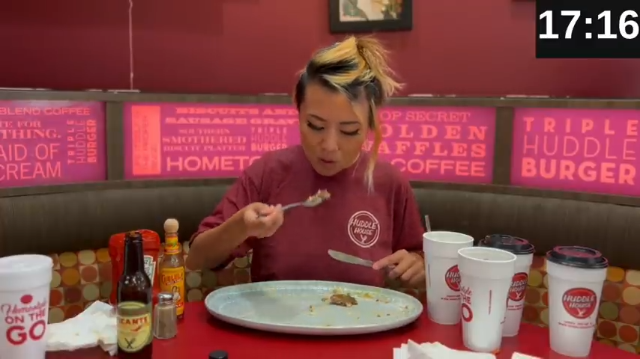} &
        \includegraphics[width=.2\linewidth]{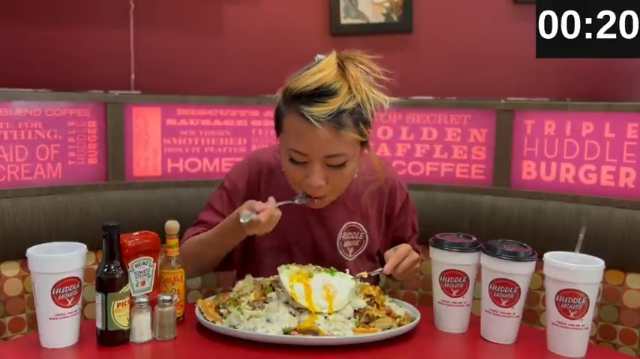} \\
        \includegraphics[width=.2\linewidth]{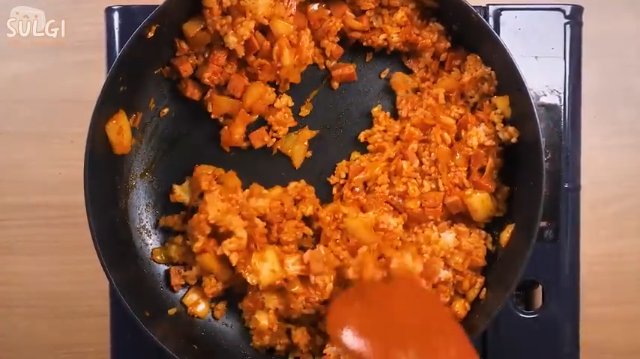} &
        \includegraphics[width=.2\linewidth]{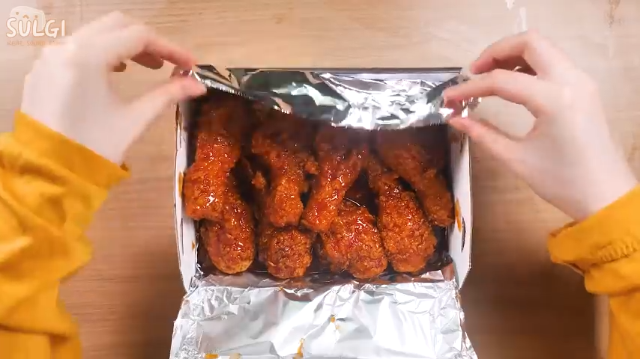} &
        \includegraphics[width=.2\linewidth]{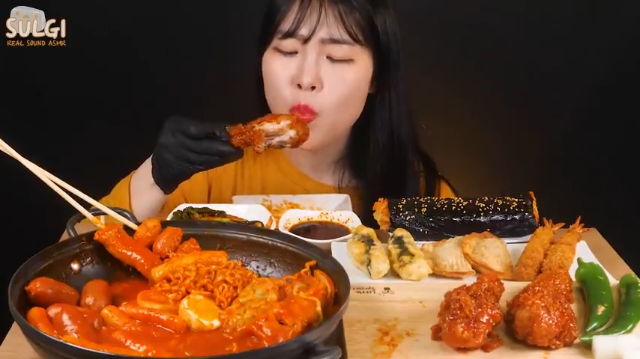} &
        \includegraphics[width=.2\linewidth]{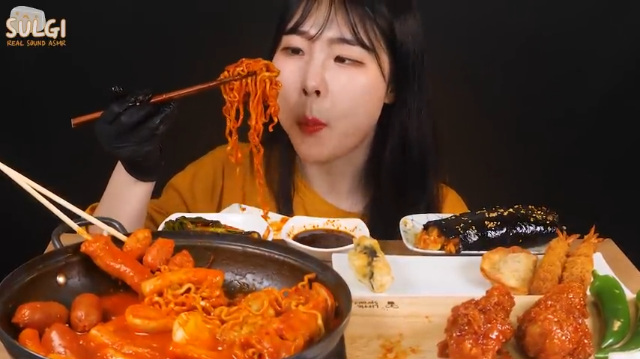} \\
        \multicolumn{4}{c}{(b) In-the-wild video dataset (YouTube)}
    \end{tabular}
    \vspace*{-1em}
    \caption{Examples of eating videos used to extract virtual IMU data using IMUTube, which include frames with zoomed shots for the food. Such frames without human motions are excluded automatically by IMUTube.}
    \label{fig:eating}
\end{figure}

\subsection{Experiment Settings and Hyperparameters}


\subsubsection{Daily Activities}

We applied 5-fold stratified cross-validation.
We could not test leave-one-user-out cross validation because the public released version of the HAD-AW dataset is incomplete, so many activities were not performed by the same users.    
After calibrating the virtual IMU data with the HAD-AW dataset, all real and virtual IMU data were segmented using a sliding window procedure with frame length of 3 seconds and step size of 1.5 seconds (identical to previous work \cite{inproceedings}).
We then extracted ECDF features \cite{hammerla2013preserving} from each analysis frame. The extracted ECDF features are then used to train a Random Forest classifier. 
We report the average of the macro F1 score from all folds for three runs. 

\subsubsection{Eating}

We used the Lab-20 dataset as training set and Wild-7 as testing set.
Accordingly, virtual IMU data was calibrated with the Lab-20 dataset.
Identical to previous eating studies \cite{thomaz2015practical}, we applied a sliding window based segmentation with a window size of 6 seconds, and 50\% overlap between consecutive analysis frames.
For each frame we then extracted mean, variance, skewness, kurtosis and root mean square features.
The predictions from the trained Random Forest (RF) model were fed into a DBSCAN clustering to aggregate eating moments (identical to previous work \cite{thomaz2015practical}).
We report binary F1-scores from the average of three runs on Wild-7 dataset.

\section{Results}

We first show overall classification performance when using virtual IMU data, as discussed earlier, for classifying activities with subtle motions. We then study how virtual IMU data impacted classification performance for each activity. 
\autoref{tab:results} shows classification results for both daily activities, and eating recognition tasks for the models trained only with real IMU data, and for those that additionally use virtual IMU data either from curated or in-the-wild dataset.


\subsection{Daily Activity Classification}

\begin{table}[t]
    \centering
    \small
    \caption{Classification accuracy for the tasks of recognizing daily activities, and eating, respectively -- evaluated through mean and binary F1 scores.
    The best performance for each task is highlighted in \textbf{bold}. 
    }
    \begin{tabular}{c||c|c}
          Task & Virtual & F1 score \\
         \hline\hline
         \multirow{3}{*}{Daily Activity~\cite{inproceedings}} 
         & - & $\textbf{0.4769} \pm 0.0008$\\
         \cline{2-3}
         & Curated & $0.4621 \pm 0.0007$ \\
         \cline{2-3}
         & In-the-wild & $0.4715 \pm 0.0011$\\
         \hline\hline
         \multirow{3}{*}{Eating~\cite{thomaz2015practical}} 
         & - & $0.7154 \pm 0.0616$\\
         \cline{2-3}
         & Curated & $\textbf{0.7994} \pm 0.0265$\\
         \cline{2-3}
         & In-the-wild & $0.7745 \pm 0.0346$\\
         \cline{2-3}
         \hline
    \end{tabular}
    \label{tab:results}
\end{table}

Using additional virtual IMU data resulted in similar classification performance when compared to the case where only real IMU data was used.
We consider that the model relied heavily on real IMU data to capture very detailed movement patterns for the classification tasks as virtual IMU data could not capture the characteristics of those subtle motions from the video data.
The result shows the \textit{breaking point} of IMUTube as using virtual IMU data no longer improves classification performance and can actually degrade model performance. We see that the model used in-the-wild virtual IMU dataset performed better than the model that used the curated dataset. This is because the in-the-wild dataset contained less data than the curated dataset, resulting in less impact to the F1 score compared to the baseline model.
In the later analysis, we will quantify failing scenario of IMUTube according to the MSI analysis for daily activities

\subsection{Eating Moment Recognition}

Much to our surprise, the addition of virtual IMU data--generated using the unmodified IMUTube system, which, again, was originally designed for the analysis of coarse motion activities--was very effective for eating specific activity classification.
Different from the daily activities recognition task, adding virtual IMU data significantly improved the performance of the downstream model.
The baseline model using only real IMU data achieved 71.5\% F1-score, and our model, trained including the curated virtual IMU data, improved the result to 79.94\%.
For the in-the-wild-scenario we observe improvements of  5.9\% to 77.45\%.
We suspect this improvement came from the virtual IMU data containing wide range of fine-grained eating motions collected from videos for varying utensils or food types that involves large arm motions.
The dataset for the daily activities classification task only included "eating a sandwich" for eating. Upon examining videos of people eating a sandwich, we find that most of the body movements involve the head. The wrist and arm movements are limited compared to other kinds of eating activities, such as eating with a fork \& knife.


\subsection{First Analysis}

The results of our initial experimental evaluation suggest that the addition of virtual IMU, generated using the vanilla IMUTube system, is only of limited effectiveness \textit{on average} for the task of recognizing activities of daily living that only exhibit subtle body (part) movements. 
However, the addition of IMUTube-generated virtual IMU data was very effective for the analysis of eating activities. 

In what follows, we dive deeper into the analysis, which resembles the aforementioned detailed ``stress test'', which will provide us with a deeper, objective understanding \textit{when}, i.e., for which activities IMUTube is beneficial, and when not. 
Aiming for concrete guidelines for practitioners who want to use IMUTube-like systems for their HAR deployments, we utilize the newly introduced MSI score as a means for quantitative assessments and guidance.

\subsection{MSI-Based Assessment of IMUTube's Utility for HAR}

\begin{figure}
    \centering
    \includegraphics[width=\linewidth]{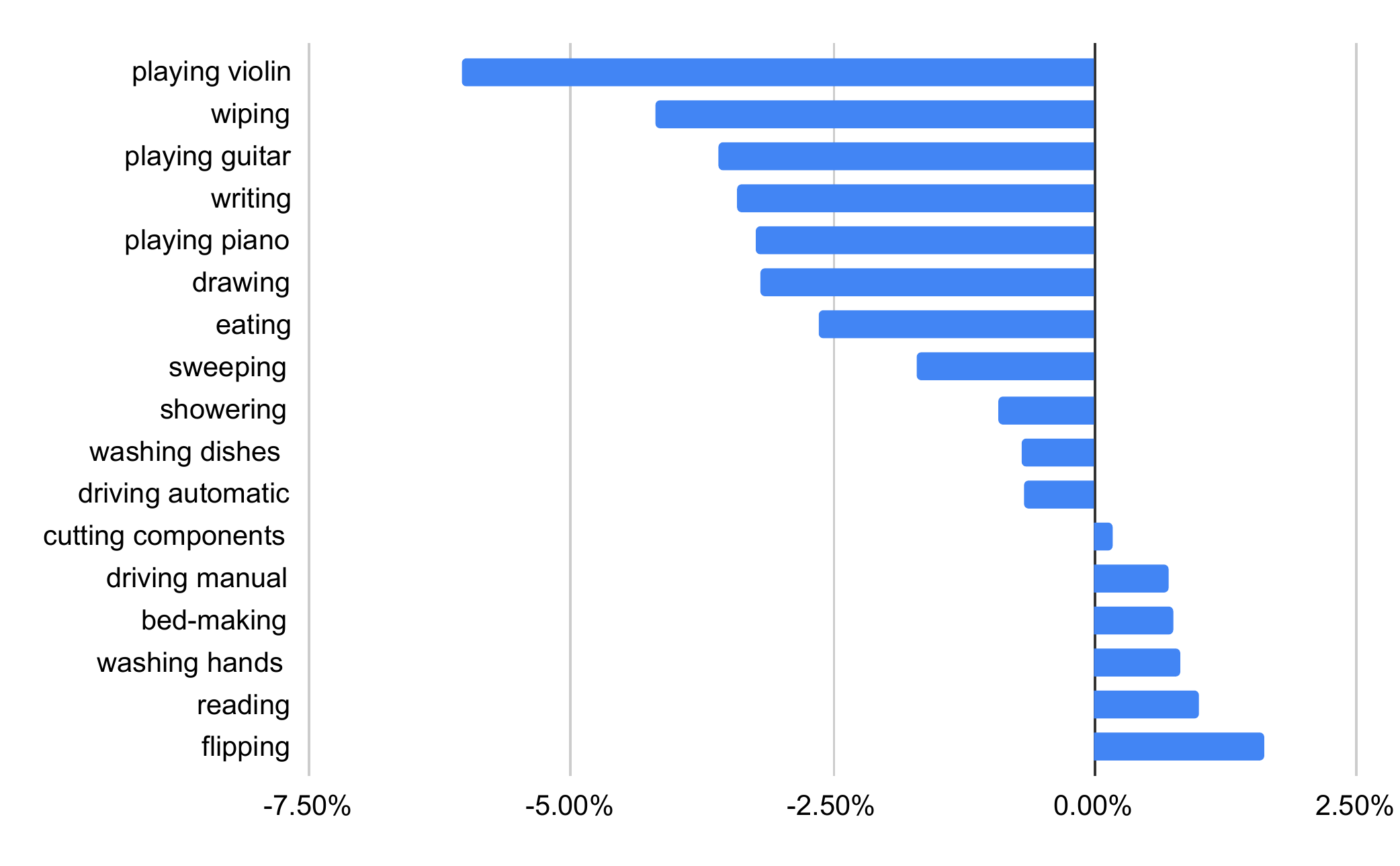}
    \vspace{-0.3in}
    \caption{F1 score change when using virtual IMU dataset additionally to real IMU data for model training. 
    The baseline model only used real IMU data for training the model.}
    \label{fig:per_class_f1}
\end{figure}

To understand for which activity virtual IMU data was most useful, \autoref{fig:per_class_f1} shows the detailed classification performance changes for each class in the daily activities recognition task when adding virtual IMU data (from curated videos) to model training.
Performance changes of up to more than $\pm 5\%$ absolute (F1 score) can be observed.
Improvements can be seen for activities like flipping, reading, washing hands, bed making, driving, and cutting.
This suggest that virtual IMU data does have value for activities with only subtle body (part) movements.

Aiming for an objective, up-front assessment--based on \textit{a-priori} analysis of the source videos--we now explore how the eventual classification performance for individual activities correlates to the video-based subtlety quantification using our MSI index.

Having a \textit{cut-off} threshold MSI value would work as a practical means for practitioners to help deciding whether or not to put resources and time into generating virtual IMU data using a system like IMUTube.
\autoref{fig:msi_cutoff} plots the changes in classification performance (``Change in F1'' -- y-axis) in relation to MSI values calculated for the underlying activity videos (``MSI'' -- x-axis).
The MSI value for each class (cMSI) represents the mode of the MSI distribution computed from all curated videos for each class using Kernel Density Estimation. 
As we increase the MSI cut-off values (x-axis), we included activity classes with more subtle classes for classification task.
Then, we computed the classification performance changes coming from using virtual IMU data at each MSI cut-off value.
For easier interpretation, we fit a spline curve (\textcolor{green}{green}) to find an approximative decision boundary between activities that benefit from additional virtual IMU data (left of the boundary), or not (right of it).

\begin{figure}
    \centering
    \includegraphics[width=0.65\linewidth]{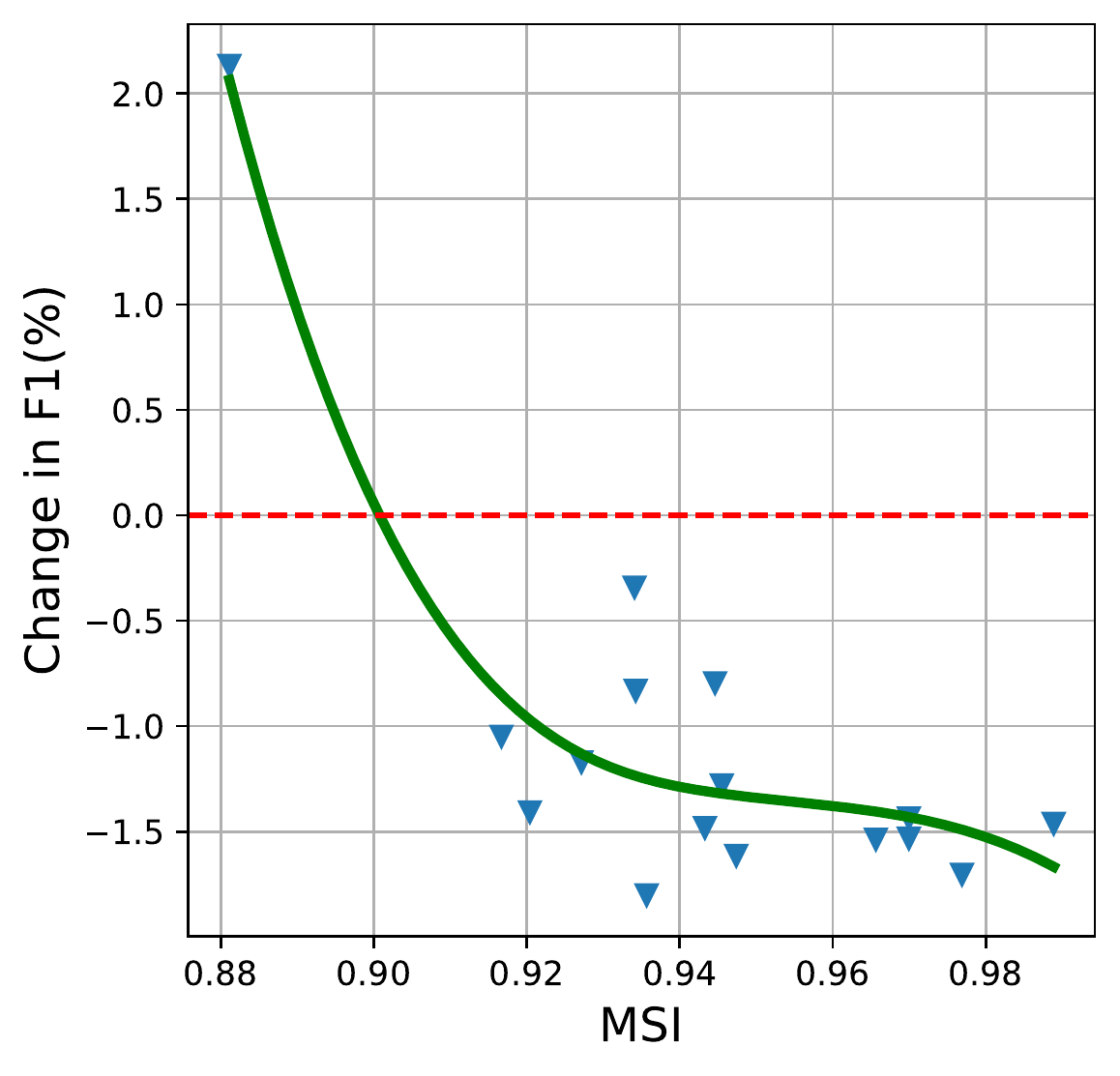}
    \caption{Classification with incrementally adding classes according to MSI cut-off values.
    We compared classification performance between the model that is trained only with real IMU data and with both real and virtual IMU data.
    The representative MSI for each class (cMSI) is the mode of MSI distribution derived from all curated videos in a class. 
    As MSI threshold (x-axis) increases, we added activity classes that has cMSI values lower than threshold values.
    We fit spline curve to find hypothetical MSI value (MSI=0.9), where model is likely not to have gains any longer by using virtual IMU data.
    }
    \label{fig:msi_cutoff}
\end{figure}

\subsection{MSI-Driven Control of Virtual IMU Data Generation}
We additionally conducted an experiment for the case of all targeting all 17 activities but generating--and using--virtual IMU data only for those classes for which the MSI value was determined to be below the cut-off value, i.e., they are predicted to be north of the decision boundary in Fig.\ \ref{fig:msi_cutoff}.
It is worth reminding that the MSI values are calculated on the videos \textit{prior} to actually generating virtual IMU data using IMUTube.

\autoref{fig:msi_cutoff_17cls} shows the strong negative correlation ($r=-0.85$, $p\leq0.001$) (Pearson) between the MSI cut-off values and changes in F1 score introduced by using virtual IMU data for the activity classes with cMSI below the MSI cut-off value.
The zero-crossing of the linear line fit for all data points was $MSI=0.89$, which was very similar to the previous experiment.
This strongly supports that MSI can provide a reference to gauge the benefit of using virtual IMU data when classifying activities with subtle motions.

Overall, this study demonstrates that the proposed MSI provides quantifiable approach to pinpoint when IMUTube will fail when it comes to subtle activities.
Activity classes beyond $0.9\leq MSI$ seems to have very subtle and complex motions in hand and wrist movements that are very difficult for IMUTube to capture due to the current limitations of state-of-the-art human motion tracking techniques~\cite{desmarais2021review}.

\section{Summary and Conclusion}

Based on the success of cross-modality transfer approaches such as IMUTube \cite{kwon2020imutube}, in this paper we explored to what extent virtual IMU data benefit HAR systems that tackle activities, which are based on more subtle body (parts) movements.
IMUTube was originally designed to support HAR systems that target coarse movement activities, such as locomotion.

Our exploration unveils two important aspects that are relevant for the broader HAR community:
\textit{i)} The subtlety of motions in activities is quantifiable in video data -- through our newly introduced Motion Subtlety Index (MSI), which correlates with the eventual downstream activity recognition accuracy on IMU data;
\textit{ii)} 
Our analysis identified a range of subtle movement activities for which IMUTube is beneficial--most notably, and somewhat surprisingly, eating-- and another range of subtle movement activities, for which it is not.

We were able to systematically, and objectively assess how the addition of virtual IMU data benefits general HAR systems and showed how the \textit{a-priori} calculation of MSI values \textit{on videos} can be used to effectively guide the application of systems like IMUTube.

In cases where the addition of IMUTube-generated virtual IMU data is not beneficial for downstream HAR accuracy, it is most likely that the underlying 3D human motion tracking techniques cannot capture enough movement detail, for example, the detailed hand and wrist coordination as it is common for subtle movements. 
As a result, the generated virtual IMU data is essentially only adding noise, which--not surprisingly--has a detrimental effect on the overall system effectiveness.

\begin{figure}
    \centering
    \includegraphics[width=0.65\linewidth]{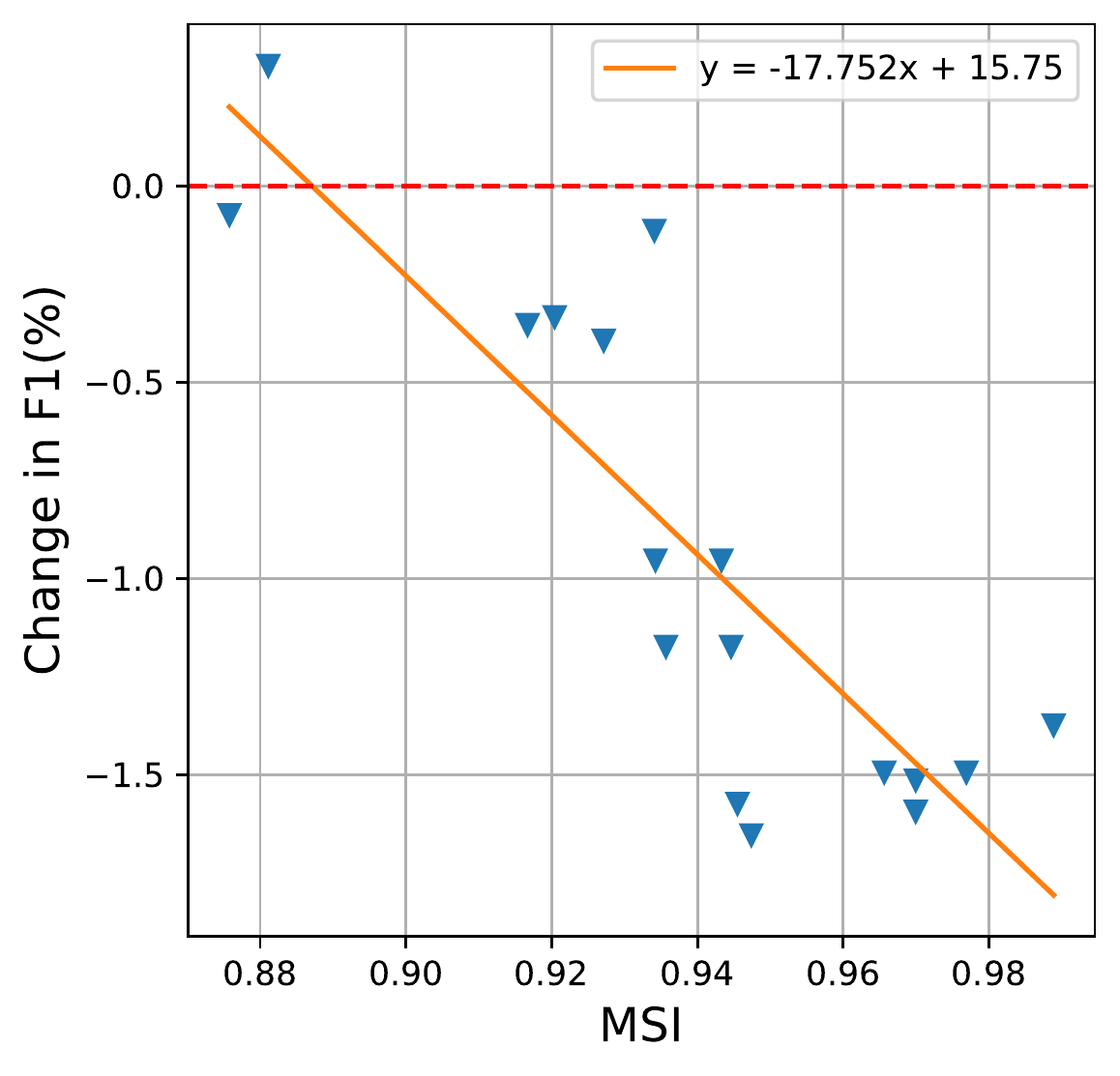}
    \caption{Recognition results for activities of daily living task where we incrementally add virtual IMU data to the training procedure only for those activities that our MSI analysis  predicted to potentially benefit from the augmentation.
    For all data points, we use all 17 activities and compare the difference in F1 score between the model that is trained only with real IMU data to the one that is based on both real and virtual IMU data.
    The representative MSI for each class (cMSI) is the mode of MSI distribution drived from all curated videos in a class. 
    As the MSI threshold (x-axis) increases, we added virtual IMU data only for activity classes that have cMSI values lower than that threshold.
    Pearson Correlation analysis shows strong negative correlation between MSI and changes in F1 score ($r=-0.85$, $p\leq0.001$).
    The zero-crossing MSI value for correlation line was $MSI=0.89$, where the model is likely to benefit any longer from adding virtual IMU data.
    }
    \label{fig:msi_cutoff_17cls}
\end{figure}

Overall, the MSI-based assessment of subtlety of activity movements and, correlated to that, the prediction of downstrem HAR effectiveness is of substantial practical value for practitioners in planning resource allocation when using cross-modality systems like IMUTube.
Such \textit{a-priori} analysis is of growing interest \cite{hiremath2020deriving} as it allows non-HAR experts to make informed decisions on what it will take to set up an effective HAR system \textit{before} actually embarking on such an endeavor. 
Finally, it is worth mentioning that the study presented in this paper is limited to wrist motions and a handful of human activities.
As such, more research and follow-up studies are encouraged for broader generalizability, specifically to: 
\textit{i)} explore the use of MSI and virtual IMU data for other on-body sensor locations and wider range of human activities that involves subtle motions; and 
\textit{ii)} to resolve the challenges in related techniques in tracking subtle motions for computer vision, graphics and biomechanics research communities.
The work presented in this paper represents the first step in this direction.


\bibliographystyle{IEEEtran}
\bibliography{
./bibs/eating,./bibs/video,
./bibs/har,./bibs/imutube,./bibs/daily_activity,
./bibs/health,./bibs/privacy}

\end{document}